\documentclass[pdf-a,balance,colorlinks,upint,subscriptcorrection,varvw,mathalfa=cal=boondoxo, spanish,french,vietnamese,russian,greek, nofoot]{asmeconf}

\hypersetup{%
	pdfauthor={},									  
	pdftitle={},                  
	pdfkeywords={},
	pdfsubject = {},			  
	pdflicenseurl={},
}

\usepackage{graphicx}
\usepackage{amsmath}
\usepackage{bbm}

\usepackage{booktabs}
\usepackage{xcolor}
\usepackage{multirow}
\definecolor{mylavender}{RGB}{160,130,210}
\newcommand{\carrie}[1]

\usepackage{siunitx}
\usepackage[capitalize]{cleveref}
\crefname{section}{Sec.}{Secs.}
\Crefname{section}{Section}{Sections}
\Crefname{table}{Table}{Tables}
\crefname{table}{Tab.}{Tabs.}

\def\confName{}

\begin{document}


\title{Bridging Design Gaps: A Parametric Data Completion Approach with Graph-Guided Diffusion Models} 
 
%
%
%

\SetAuthors{%
	Rui Zhou\affil{1}\CorrespondingAuthor{}, 
	Chenyang Yuan \affil{2}, 
	Frank Permenter\affil{2}, 
        Yanxia Zhang\affil{2}, 
        Nikos Arechiga\affil{2},
        Matt Klenk\affil{2},
	Faez Ahmed\affil{1}  
	\CorrespondingAuthor{\{zhourui, faez\}@mit.edu}
	}

\SetAffiliation{1}{Massachusetts Institute of Technology, Cambridge, MA }
\SetAffiliation{2}{Toyota Research Institute, Los Altos, CA}


\maketitle
\bibliographystyle{unsrt}



\begin{abstract}
\begin{figure*}[!hbt]
\centering
\includegraphics[width=13cm]{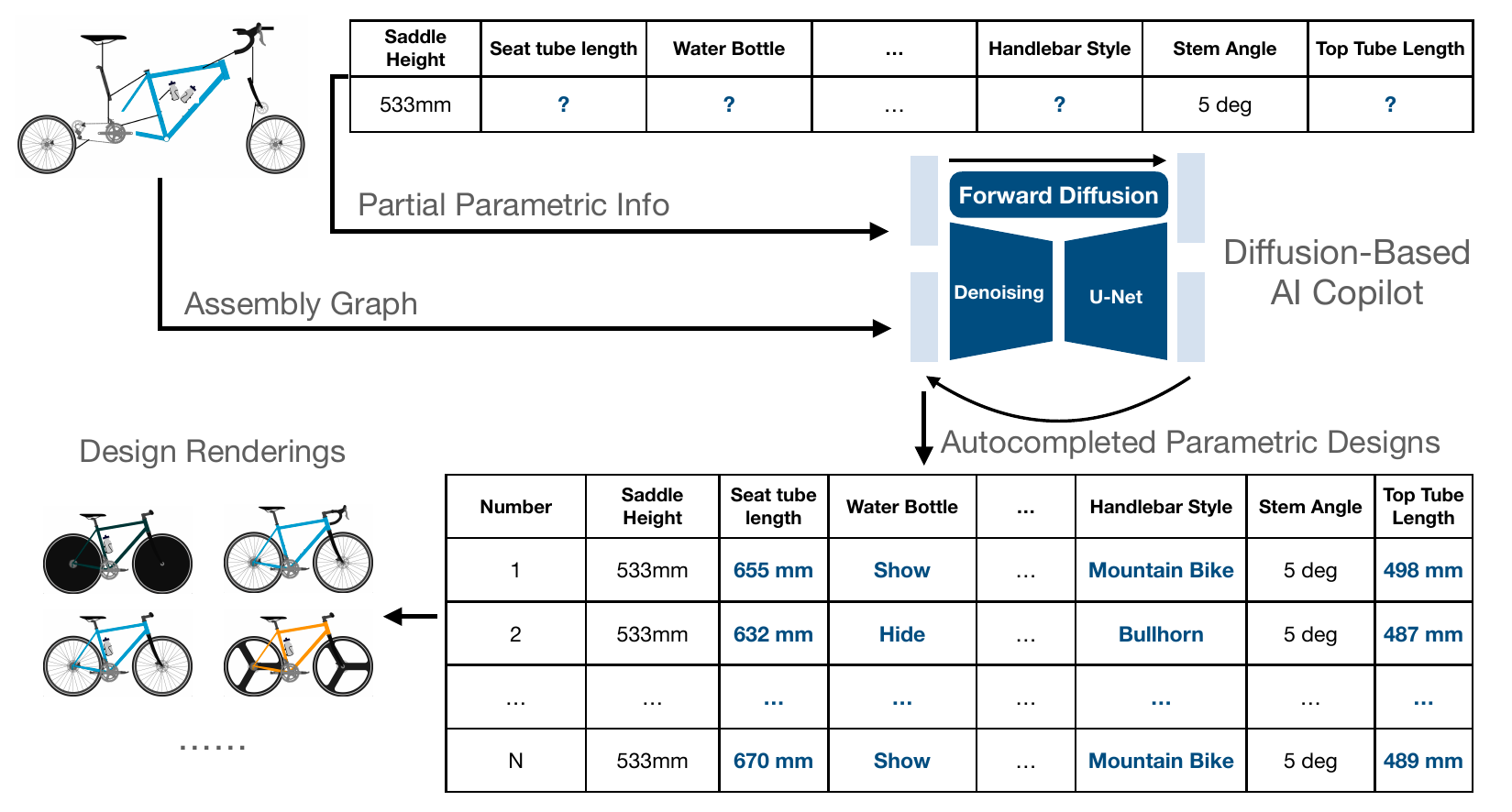}
\caption{Diagram showcasing our model's capability as an AI copilot that autocompletes the engineering designs based on incomplete user-defined information.}\label{fig:overview}
\end{figure*}

This study introduces a generative imputation model leveraging graph attention networks and tabular diffusion models for completing missing parametric data in engineering designs. This model functions as an AI design co-pilot, providing multiple design options for incomplete designs, which we demonstrate using the bicycle design CAD dataset. Through comparative evaluations, we demonstrate that our model significantly outperforms existing classical methods, such as MissForest, hotDeck, PPCA, and tabular generative method TabCSDI in both the accuracy and diversity of imputation options. Generative modeling also enables a broader exploration of design possibilities, thereby enhancing design decision-making by allowing engineers to explore a variety of design completions. The graph model combines GNNs with the structural information contained in assembly graphs, enabling the model to understand and predict the complex interdependencies between different design parameters. 
The graph model helps accurately capture and impute complex parametric interdependencies from an assembly graph, which is key for design problems. By learning from an existing dataset of designs, the imputation capability allows the model to act as an intelligent assistant that autocompletes CAD designs based on user-defined partial parametric design, effectively bridging the gap between ideation and realization.
The proposed work provides a pathway to not only facilitate informed design decisions but also promote creative exploration in design.

\end{abstract}

\section{Introduction}

\label{sec:intro}
    The advent of generative models has ushered in a transformative era in engineering design, offering many opportunities for automation, customization, and efficiency. Despite their potential, these models predominantly focus on generating new samples from the ground up, overlooking their application in addressing incomplete or missing data---a frequent scenario in engineering.
     Given the complex dependency of engineering designs on each parameter, missing data significantly hampers the generation of effective designs. To bridge this gap, our study introduces an innovative approach employing diffusion models and Graph Neural Networks (GNN), combined with assembly graphs for feature encoding. 
    This work not only aims to redefine how missing data is interpreted and completed in the context of engineering designs but also views generative imputation tasks as a design recommendation system, where a user with an incomplete design gets multiple recommendations on how to complete the design.

    Assembly graphs, which detail the hierarchical and spatial relationships among various components of a design, provide a rich source of structural information that is typically underutilized in traditional imputation methods. By embedding this structural insight into a GNN framework, our model captures the nuanced interdependencies between design parameters, offering a more refined and contextually aware approach to data imputation.

     Traditional imputation techniques often fall short in these scenarios \cite{review}, either oversimplifying the complexity of the design parameters or failing to account for their interrelated nature. As shown in Figure \ref{fig:overview}, our GNN-based model addresses these shortcomings, leveraging the detailed guidance provided by assembly graphs to inform its predictions and fill in the gaps with a level of precision previously unattainable.

    Beyond its immediate utility in completing parameter sets, the model's integration into a larger generative design framework marks a pivotal advancement. It enables the model to function as a copilot in the design process, where it not only fills in missing information but also suggests design alterations and improvements based on the comprehensive understanding it has of the design space. This capability transforms the model from a passive tool into an active participant in the design process, opening up new avenues for collaborative design between human engineers and AI systems.

    This research paper contributes to the field of engineering design and generative modeling with the following contributions: \begin{enumerate}
    \item \textbf{Innovative Diffusion Model Architecture:} We present an innovative diffusion model architecture that explicitly incorporates parameter hierarchy through assembly graphs. This approach is novel in its consideration of the structural and relational intricacies of design components, leading to more accurate imputation of missing parametric data. This architecture's efficacy is demonstrated through its application to tabular data, showcasing its versatility and potential for broader applications beyond the specific case of bike CAD generation.

    \item \textbf{Comparative Performance Evaluation:} Through extensive comparative evaluations against three classical imputation methods, our model exhibits enhanced performance in terms of both accuracy and diversity of generated design options. Further, by comparing against a state-of-the-art diffusion imputation model, we show that the ability to encode information represented by the assembly graphs increase the diffusion model's performance significantly.

        \item \textbf{Generative Imputation for Design Recommendations:} We reimagine generative imputation task as a means for providing design recommendations. By using bike CAD design as a practical example, we demonstrate how our model acts as an AI design copilot, offering diverse design completions based on incomplete parametric input for both continuous and categorical variables.
    \end{enumerate}

\subsection{Related Work}

The following sections delve into the evolution of data imputation methods, starting from classical statistical techniques to the advent of deep learning approaches, highlighting their respective contributions and limitations in addressing missing data challenges.

\label{sec:related_work}
\subsubsection{Classical Imputation Methods}
   Classical statistical techniques, such as mean or median imputation and regression-based imputation, have been popular methods for data imputation tasks. While mean/median imputation offers simplicity, it often introduces bias and fails to capture the underlying data complexity \cite{little2019statistical}. Since the advent of machine learning, researchers have sought to use classical machine learning methods to achieve better results. For example, regression-based imputation, utilizes relationships between variables to predict missing values \cite{schafer1997analysis}. However, it may not adequately handle non-linearities or complex interactions between features.  Earlier works focus on using classical machine learning methods. For example, Troyanskaya et al. employ a K-Nearest Neighbors (KNN) approach using similar instances \cite{troyanskaya2001missing}, offering a more nuanced approach to imputation. Stekhoven et al. use Random Forests and decision trees to address non-linearities and feature interactions effectively in \cite{stekhoven2012missforest}. In \cite{mazumder2010spectral}, Mazumder et al. use Matrix Factorization techniques to learn latent feature representations to handle missing data in matrices. 
   However, these classical approaches often work better in low dimensional spaces, and assume that data is missing at random, which may not always hold true, especially in complex datasets. 
\subsubsection{Deep Learning Based Imputation Methods}
    The integration of deep learning into the field of data imputation has marked a pivotal shift toward addressing the complex and nuanced nature of missing data. Unlike traditional statistical methods that often rely on assumptions about data distribution or the relationships between variables, deep learning approaches leverage the ability to model data distributions in high-dimensional spaces, enabling the capture of intricate patterns and dependencies that are not immediately apparent. More advanced methods, such as Generative Adversarial Networks (GANs) \cite{gan} and Variational Autoencoders (VAEs) \cite{vae}, have been shown to introduce significant improvements for data imputation tasks. For example, Multiple imputation using denoising autoencoders (MIDA) \cite{gondara2018mida} utilizes denoising autoencoders to learn robust data representations through training on corrupted data, facilitating the accurate reconstruction of missing entries. In \cite{chen2019traffic}, Chen et al. introduced a novel approach for traffic flow imputation that leverages parallel data and Generative Adversarial Networks (GANs). Another case is the GAIN model introduced in \cite{gain}, Yoon et al. utilize a GAN architecture to significantly improve the performance of imputation tasks. In \cite{fortuin2020gp}, Fortuin et al. introduce imputation of missing values in multivariate time series data, leveraging a deep probabilistic approach with Gaussian processes within a VAE framework. 
    Further, in the design domain, many works have utilized deep learning-based methods to handle complex design tasks. For example, In \cite{jmd1}, Zhu et al. utilized pre-trained transformers to create bio-inspired designs. In \cite{jmd2}, Raina et al. designed a novel learning-based architecture for learning from human designers. Regenwetter et al. used an AutoML approach for structural performance analysis of bike frames and Song et al. designed an architecture allowing multimodal learning for conceptual design evaluations in \cite{song2023attention}. However, there are very few works on data imputation for engineering design problems, where the model needs to consider both design information and engineering information. 
Our work addresses this gap by reimagining data imputation as a design recommendation system, focusing on the unique challenges of part-assembly hierarchies in engineering designs. We enhance a state-of-the-art diffusion model to specifically cater to these challenges, aiming to provide a more intuitive and effective solution for engineering design imputation.

\subsubsection{Diffusion and Graph-based Imputation Methods}
\begin{figure}[ht]
\centering
\includegraphics[width=8cm]{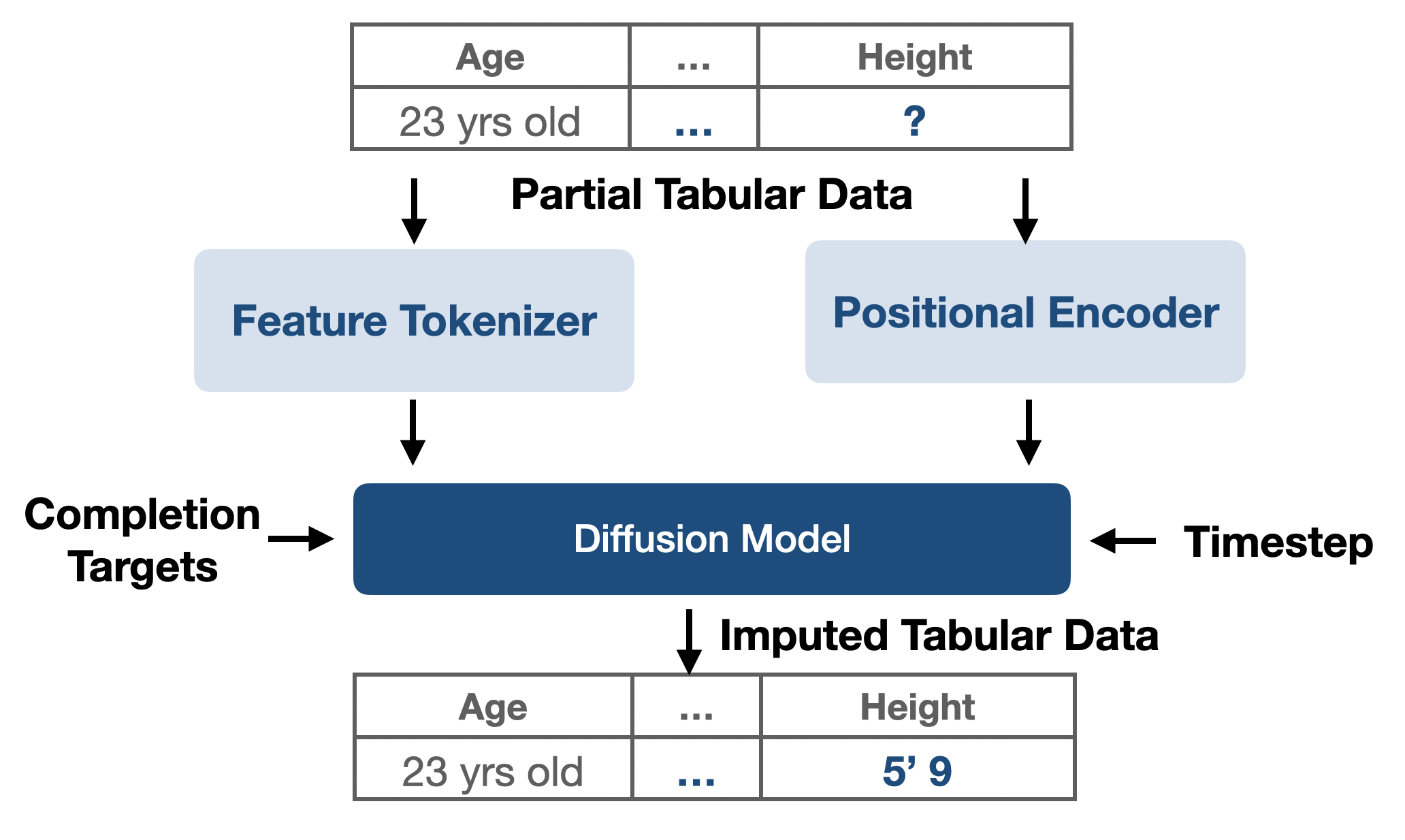}
\caption{Diagram shows TabCSDI's model architecture.}\label{fig:tabcsdi}
\end{figure}
    The introduction of diffusion models in \cite{diffusion} has brought exciting new possibilities to the data imputation field. This has been exemplified by \cite{csdi}, where Tashiro et al. introduced Conditional Score-based Diffusion models for Imputation (CSDI), an approach for probabilistic time series imputation that leverages score-based diffusion models conditioned on observed data, significantly enhancing imputation accuracy in healthcare and environmental datasets. The authors show that this method not only outperforms existing probabilistic and deterministic imputation techniques by a substantial margin but also demonstrates its versatility in applications such as time series interpolation and probabilistic forecasting. As an improvement to the CSDI model, Zheng and Charoenphakdee introduced Conditional Score-based Diffusion Models for Tabular data (TabCSDI) in \cite{tabcsdi}. 
    We show TabCSDI's architecture in Figure \ref{fig:tabcsdi}. TabCSDI uses a feature encoder and position encoder to encode tabular data and then uses a diffusion model to impute the missing data. It effectively handles both categorical and numerical variables through innovative encoding techniques, and demonstrates superior performance over existing methods on benchmark datasets, highlighting the potential of diffusion models in addressing the challenges of tabular data imputation.

    Another significant innovation is the introduction of Graph Neural Networks (GNNs) in \cite{scarselli2008graph} and Graph Convolutional Networks (GCNs) in \cite{kipf2016semi}. It opens new opportunities for many domains, especially in engineering design, where graphs can represent relationships between different items or parameters. GNNs have also been used for data imputation, for example, in \cite{spinelli2020missing}, Spinelli et al. introduced an approach to missing data imputation using adversarially-trained GCNs, formulating the task within a graph-denoising autoencoder framework. Their method, which leverages the similarity between data patterns encoded as graph edges, demonstrates superior performance over traditional imputation methods across various datasets, showcasing the potential of GCNs in enhancing the accuracy and robustness of missing data imputation. Furthermore,  You et al. (2020) introduced GRAPE in \cite{you2020handling}, a graph-based framework for addressing missing data through graph representation learning, where data observations and features are conceptualized as nodes in a bipartite graph. This innovative approach, which applies Graph Neural Networks for both feature imputation and label prediction, demonstrates significant improvements over traditional methods, offering a promising new direction for handling missing data in machine learning tasks. Further, new GNN architectures introduce new ooportunies for improving the performance of existing models. For example, in \cite{GAT}, Velickovic et al. introduced Graph Attention Networks (GAT) that utilize self-attention mechanisms to overcome the shortcomings of traditional GNNs. To further improve the performance, \cite{GATV2} introduced the next generation of GATs named GATv2, which has dynamic attention mechanisms and is more expressive that GATs. However, graph neural networks have not been combined with diffusion models for data imputation tasks, nor have they been applied to engineering problems for data imputation.
    \\
    In this work, we seek to combine the advantages of both diffusion models generative modeling capabilities and GCNs relationship modeling capabilities to achieve higher performance for complex engineering design tasks. 
    Specifically, we are interested in enabling the use case where the model can act as an AI agent that autocompletes the engineering design with the prompting of partial parametric information.

\section{Methodology}
\begin{figure*}[!hbt]
    \centering
    \includegraphics[width=14cm]{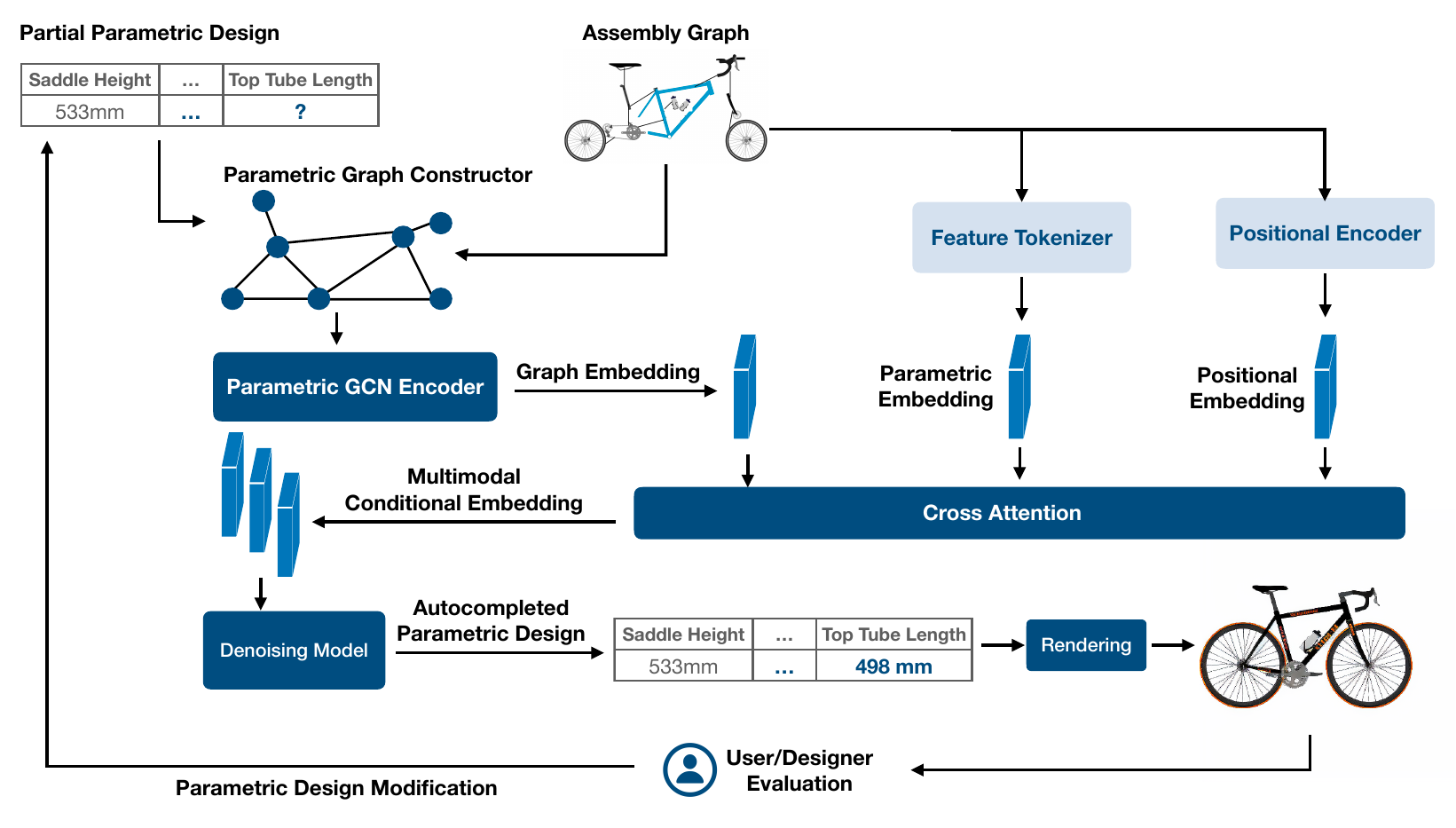}
    \caption{Overview of our model.} 
    \label{fig:overview}
\end{figure*}
\label{sec: method}
\subsection{Problem Formulation}
Specifically, we are interested in the problem of generating the completed parametric product assembly design (e.g., a bicycle) conditioned on the partial parametric information defined by the user.\\
Specifically, we are interested in the following problem: We denote a complete parametric design for a product assembly as $X_{complete}=\{x_{1:D}\}\in\mathbb{R}^D$, where D is the total number of features for a complete product assembly design in parametric design form. Thus, we represent a partial parametric design for a product as $X_{partial}=\{x_{1:D}\}\in(\mathbb{R\cup \varnothing})^D$, where each $x_i$ can be either missing, categorical, or numerical. If $x_i$ is categorical, then its valid ranges are $c_1,.., c_C$, where $c_i$ denotes the categories and C is the total number of categories for this feature. Thus, our goal is to find $F_{autocomplete}: X_{partial}\rightarrow X_{full}$, where $F_{autocomplete}$ is our model. Further, we note that the diversity in the autocompleted designs is also extremely important to realize the goal of an engineering design copilot that can autocomplete partial designs. Thus, we are interested in finding:

\begin{equation}
    P_{missing|Observed}=P(m_1, ..., o_M \in M | o_1, ..., o_O \in O),
\end{equation} where $M$, the set of missing features, is denoted by $M=\{ m_i \in X_{partial}: m_i = \varnothing\}$, and $O$, the set of observed or defined parameters, is denoted by $O=\{ o_i \in X_{partial}: o_i \in \mathbb{R}\}$. 
To measure the performance of the task, we are interested in three main aspects:
\begin{enumerate}
    \item Accuracy for numerical features:
    Following TabCSDI in\cite{tabcsdi} and CSDI in \cite{csdi}, we use Root Mean Square Error (RMSE) to measure the models' performance on numerical features. That is, for numerical features, we calculate the RMSE as follows:
        \begin{equation}
        \begin{aligned}
            &RMSE(X_{test})=
            &\frac{1}{N_{test}}\sum_{i=1}^{N_{test}}\sqrt{\frac{\sum_{j=1}^{M_i}(\hat{x}_i^j-y_i^j)^2}{M_i}},
        \end{aligned}
    \end{equation} where $N_{test}$ is the number of test samples, $M_i$ is the number of missing numerical features for sample $i$, $\hat{x}_i^j$ is the models' imputation prediction for missing feature $j$ of sample $i$, and $y_i^j$ is the dataset value for missing feature $j$ of sample $i$, which we use as the reference target design for this testings sample.
    \item Accuracy for categorical features: To compare the model performance in imputing categorical features, we use Error Rates~\cite{tabcsdi}. That is, for categorical features, we use:
    \begin{equation}
        \begin{aligned}
            &Err(X_{test})=\\
            &\frac{1}{N_{test}}\frac{1}{M_{i}}
            \sum_{i=1}^{N_{test}}\sum_{j=1}^{M_{i}} \mathbbm{1} [\hat{x}_i^j\neq y_i^j],
        \end{aligned}
    \end{equation}, where $N_{test}$ is the number of test samples, $M_i$ is the number of missing categorical features for sample $i$, and $\mathbbm{1}$ is an indicator function. $y_i^j$ is the dataset value for missing feature $j$ of sample $i$, which we use as the reference target design for this testings sample.
    \item Diversity: Unlike other imputation tasks, design recommendation by an AI design copilot necessitates that the set of designs obtained by a generative imputation model are diverse. 
    Hence, we add a third metric--diversity--to compare different imputation models. Specifically, we use three metrics to measure the performance of the models: (a) Diversity Score to measure coverage, and (b) KL-Divergence of Generated Features' Distributions when compared to those of the dataset to measure representativeness~\cite{regenwetter2023beyond}. Specifically, we measure the Diversity Score as Follows:
    \begin{equation}
        \begin{aligned}
            &Diversity(X_{test})=\\
            &\frac{1}{N_{test}}\frac{1}{M_{i}}
            \sum_{i=1}^{N_{test}}\sum_{j=1}^{M_{i}} max_{k\neq l }\sqrt{(S_{k,j}-S_{l,j})^2},
        \end{aligned}
    \end{equation}, where $l$ and $k$ are the index of the samples obtained by the generative model and $S_{k,j}$ and $S_{l,j}$ are the $k$-th and $jl$-th sampled value for missing feature $j$. The Diversity Score finds the average maximum distance between missing features. In our problem, we set the number of samples to 50 to balance evaluation speed and evaluation accuracy.

\end{enumerate}

\paragraph{Interpreting metrics for generative data imputation:}
Evaluating the performance of generative data imputation models involves a nuanced understanding of the data's conditional distributions. For instance, consider two scenarios where different attributes are missing: the length of the top tube in one and the color of the bike in another. The length of the top tube is constrained by specific geometrical principles, implying a single correct value that maintains the integrity of the bike's structure. On the other hand, the color of the bike, being an attribute with no technical constraints, can vary widely, allowing for a multitude of correct possibilities. In such a context, for the first scenario, an imputation model's success is measured by its accuracy and its ability to converge on the singular correct value, reflecting low diversity. Conversely, for the bike's color, the model's ability to offer a wide range of plausible options indicates success, showing high diversity even if individual predictions vary widely. This illustrates that not all missing features are created equal; some will have tightly constrained correct values, while others will naturally allow for greater variation. Thus, when calculating the diversity score in our problem, we only consider features that have a mean correlation value that is greater than the median among all features of the dataset. Therefore, we adopt a comprehensive approach in reporting our findings, acknowledging the diverse nature of data and the varying benchmarks for what constitutes good performance in generative data imputation.

\subsection{Overall Pipeline}

Our methodology introduces a new pipeline to tackle missing parametric data in engineering design. It consists of five steps that combine Graph Neural Networks (GNNs) with advanced machine learning for accurate data imputation and visualization.
Initially, (1) the pipeline accepts partial parametric design and an assembly graph, utilizing these inputs to construct a detailed graph that accurately describes the features and their interdependencies within the design. Following this, (2) it employs feature and positional tokenizers from TabCSDI to translate the parametric and spatial information into a tokenized format. The third step, (3) involves the application of cross-modal attention to fuse the multimodal conditional embeddings derived from the graph, alongside the outputs of the feature and positional tokenizers. Subsequently, (4) a diffusion-based denoising model is utilized to accurately impute the missing features. Finally, (5) the pipeline forwards the now-complete parametric design to a CAD rendering engine, which generates an image of the fully realized design that can be shown to an end-user. 
This end-to-end process not only addresses the imputation of incomplete data but also facilitates the generation of accurate and detailed engineering designs, bridging the gap between partial information and complete design realization.
\subsection{Construction of Graphs}

\begin{table*}[!ht]
  \centering
  \caption{Categorization of features into bike components}
    \begin{tabular}{lp{34em}}
    \toprule
   {\textbf{Component Name}} & \multicolumn{1}{l}{{\textbf{Features}}} \\
    \midrule
    Seat Tube & Seat tube length, Seat tube extension2, Seat tube diameter, Seatpost setback, Seatpost LENGTH, Stack, Lower stack height, Upper stack height \\
    Head Tube & Head tube upper extension2, Head angle, Head tube lower extension2, Head tube diameter \\
    Top Tube & Top tube rear diameter, Top tube front diameter \\
    Down Tube & Down tube front diameter, Down tube rear diameter \\
    Chain Stay & CHAINSTAYAUXrearDIAMETER, Chain stay horizontal diameter, Chain stay position on BB, Chain stay taper, Chain stay back diameter, Chain stay vertical diameter \\
    Seat Stay & Seat stay junction0, Seat stay bottom diameter, SEATSTAY\_HF, SEATSTAY\_HR, SEATSTAYTAPERLENGTH \\
    Fork  & FORK0R, FORK0L \\
    Saddle & Saddle P, Saddle thickness, Saddle angle, Saddle J, Saddle H, Saddle E, SADDLETIPtoMIDDLE, Saddle length \\
    Wheel & Wheel width rear, Wheel width front, Wheel diameter front, Wheel diameter rear, SPOKES composite front, SPOKES front, SPOKES rear, SPOKES composite rear, ERD rear, ERD front \\
    Handle & Road bar reach, Road bar drop, Brake lever position, Bullhorn angle, HBAREXTEND, Handlebar angle, MtnBar angle, HBARTHETA, Pedal width, Stem angle, Stem length \\
    BB    & BB textfield, BB length, BB diameter \\
    \bottomrule
    \end{tabular}%
  \label{tab:component features}%
\end{table*}%
\begin{figure}[t]
\centering
\includegraphics[width=8cm]{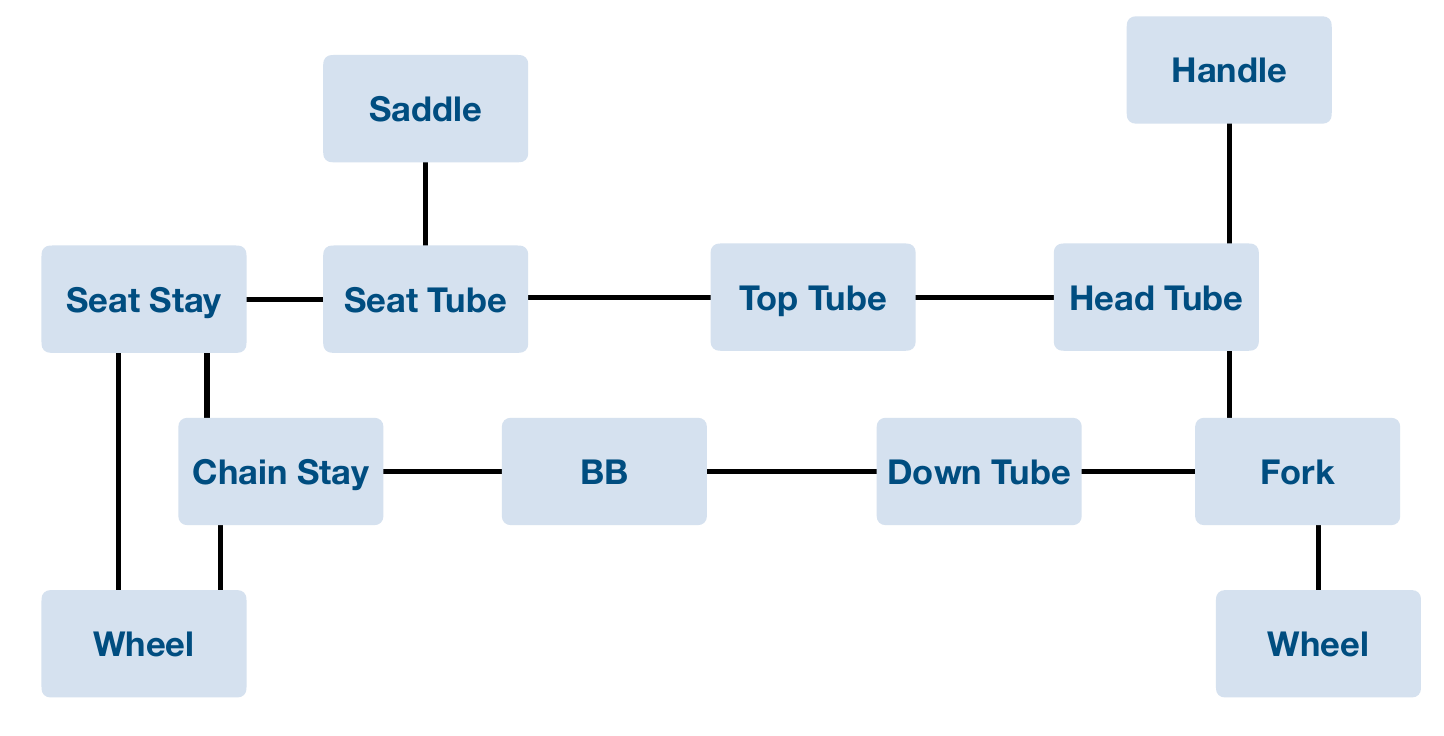}
\caption{Structure of bike assembly graph.}\label{fig:bike_graph}
\end{figure}
We use an abridged version of the BIKED dataset~\cite{10.1115/1.4052585} for this work, where each bike is represented by 221 features. 
We categorize these features into 11 distinct components, as outlined in Table \ref{tab:component features}. This classification forms the foundation for defining the assembly graph, visually represented in Figure \ref{fig:bike_graph}, which delineates the structural and relational framework of the bike components. Utilizing this predefined assembly graph, we then systematically construct feature-specific graphs for each data input. These graphs serve as the input to the GCN, enabling context-aware encoding of the features based on their interconnections and the overall design architecture. This approach ensures that the GCN is informed by a detailed understanding of the bike's structure, facilitating accurate feature imputation and encoding within the pipeline.
\subsection{Assembly Graph Encoding with GCNs}
Here, node features within the graph are constituted by the concatenated features of each component. Furthermore, for features that are missing and to be imputed, we set their values to 0 to preserve the same feature sizes. An edge is established between nodes if their corresponding components are physically connected or interact within the bike's structure, reflecting the assembly graph's delineation of component relationships. This encoding strategy allows the GCN to process the bike design in a manner that respects the inherent connectivity and dependency between different parts. By embedding the assembly graph's relational information into the GCN, we enable the network to accurately capture and encode the complex interplay of bike components, facilitating a more nuanced and informed imputation and subsequent generation process.
\subsection{Multimodal Encoding}

\begin{table*}[!htbp]
\centering
\caption{Comparison with deterministic models indicating performance metrics, where lower is better ($\downarrow$) and higher is better ($\uparrow$). Boldface denotes the best performance in each metric.}
\label{table:deterministic_models}
\sisetup{table-format=2.2} 
\begin{tabular}{
  l
  S
  S
  S
  S[table-format=2.2] 
}
\toprule
{Metric} & {Ours (best)} & {PPCA} & {HotDeck} & {MissForest} \\
\midrule
RMSE $\downarrow$        & 0.92 & 1.26  & 36.99 & \textbf{0.65} \\
Error Rates $\downarrow$ & \textbf{0.18} & {N/A} & 0.66  & 0.99  \\
Diversity Scores $\uparrow$ & \textbf{3.10}  & 0.00   & 0.00    & 0.00   \\
\bottomrule
\end{tabular}
\end{table*}

We draw inspiration from the TabCSDI framework \cite{tabcsdi} for the encoding process that models both the features and their positional context within the design space. To achieve this, we incorporate a Feature Tokenizer and a Positional Tokenizer. The Feature Tokenizer employs fully connected layers to process and encode numerical features, alongside a dedicated embedding layer designed specifically for the categorical features, ensuring that each feature type is optimally represented. Concurrently, the Positional Tokenizer addresses the spatial aspect of the data, encoding the positional information of features to highlight the inherent relational context found in tabular data, where features positioned closely are likely to convey related information.
To fuse the multimodal information, we use a cross-attention mechanism. This mechanism fuses the embeddings derived from the assembly graph with those generated by the Feature and Positional Tokenizers, effectively integrating the structural, categorical, and spatial dimensions of the data into a singular multimodal conditional embedding. This integrated embedding captures both the intrinsic properties and interrelations of features within the assembly graph.

\subsection{Full Parametric Design Generation and Rendering}
We feed the multimodal embedding vector—enriched with assembly graph, feature, and positional embeddings—into a diffusion-based denoising model. This model, tasked with autocompletion, generates the missing parameters, leveraging the diffusion process to ensure both the diversity and accuracy of the parametric designs. The generated complete parametric design is then processed through a rendering pipeline, transforming it into the final design visualized in image format. 

\section{Experiment Results}
\subsection{Dataset}
We base our problem on the BIKED Dataset introduced by Regenwetter et al. in \cite{10.1115/1.4052585}, which contains the CAD files and parametric information of 4500 individually designed bicycles. We augment the dataset by randomly sampling the valid ranges for existing features to create an augmented version of the dataset, resulting in 12,506 samples. We further do a training test split with a 90-10 ratio. Similar to TabCSDI~\cite{tabcsdi}, we randomly mask out 10\% of the features for the model to impute. We create the testing set by randomly masking out 10\% of the features as well. For overall accuracy metrics such as RMSE and Error Rates, we report the results on the testing dataset which has 10\% of features masked out. For feature-specific studies, we report the results by only masking out the specific feature being studied. For repeatability, we generate 50 samples for each test case. 

\begin{table*}[!htbp]
\centering
\caption{Comparisons with generative models, highlighting performance across various metrics. Lower values are better ($\downarrow$), except for the Diversity Score where higher is better ($\uparrow$). Boldface indicates the best performance in each category.}
\label{table:gen}
\sisetup{table-format=1.3} 
\begin{tabular}{
  l
  S
  S
  S
  S
  S
}
\toprule
{Metric} & {RMSE $\downarrow$} & {$R^2$ $\downarrow$} & {MAE $\downarrow$} & {Error Rate $\downarrow$} & {Diversity Score $\uparrow$} \\
\midrule
Ours (GATV2)         & \textbf{0.916} & \textbf{0.161} & \textbf{0.335} & \textbf{0.184} & \textbf{3.10} \\
Ours (Vanilla GCN)   & 0.924           & 0.146           & 0.349           & 0.19            & 2.16           \\
TabCSDI              & 0.950           & 0.098           & 0.36            & 0.213           & 1.47           \\
\bottomrule
\end{tabular}
\end{table*}
In our study, we benchmark our models against state-of-the-art diffusion model TabCSDI \cite{tabcsdi} as well as popular classical models, such as MissForest~\cite{stekhoven2012missforest}, hotDeck~\cite{hotdeck}, and PPCA~\cite{qu2009ppca}. We divide the comparisons into two groups: 
\begin{enumerate}
    \item Comparisons against popular deterministic models: We show that our model achieves better or comparable results while generating diverse designs, rather than a fixed design. Unlike generative models, classical models generate a single imputed value, which is less useful for design exploration and recommendation.
    \item Comparison of GCN choices against the state-of-the-art generative model, TabCSDI in \cite{tabcsdi}: We show that our models have significantly more accurate design generations than the state-of-the-art model in tabular data imputation. 
Through these two sets of quantitative evaluations, we show that our model achieves superior performance in both groups and thus can generate engineering designs that are both diverse and accurate. 
\end{enumerate}
\subsection{Comparison with deterministic models}
Our model was compared to popular deterministic models, including MissForest and hotDeck, as well as PPCA. Here we test the models on the testing dataset with 10\% of the features randomly masked out. The masked-out features are different for each testing case.  We show the results in Table \ref{table:deterministic_models}. Our model marked as "Ours (best)" achieves the lowest Error Rates on categorical features and comparable RMSE on numerical features while being able to generate diverse outputs. Our model achieves a diversity score of 3.10, which indicates our model is able to generate engineering designs with higher variety while ensuring accuracy. The comparison revealed that our model not only achieved better or comparable results in terms of accuracy but also excelled in producing diverse designs. Unlike the deterministic models that typically yield a single, fixed output, our approach enables the generation of multiple design variations from the same set of initial conditions. This capability is pivotal for engineering applications where design flexibility and exploration are essential. 

\subsection{Comparisons with generative models}
In a focused comparison with TabCSDI, our model demonstrated a significant advantage in design generation accuracy. We further compare the results achieved using different choices of GCNs. We show the results in Table \ref{table:gen}. We show that our models achieve the lowest values on RMSE, $R^2$, MAE, and Error Rates. Further, we show that using GATV2 achieves better performance than using a Vanilla GCN as the graph encoder. 
This could be explained by the attention networks providing better modeling capabilities for graphs with high-dimensional features.
Our model's use of Graph Attention Networks informed by assembly graphs provided a more structured and context-aware approach to imputation, resulting in designs that more accurately reflect real-world constraints and specifications.

\subsection{Feature Distributions Analysis}
In this section, we study the model's output for two types of distributions: 
\begin{enumerate}
    \item Overall distributions: We compare the model's generated feature's distributions against those of the dataset, demonstrating that our model is capable of generating feature distributions very close to the original one.
    \item Conditional Distributions: We study the models generated features' distributions for specific bikes. We show that our model generates accurate conditional features for features of varying levels of correlations with other features. 
\end{enumerate}
\subsubsection{Overall Distributions}
\begin{figure}[t]
\centering
\includegraphics[width=8cm]{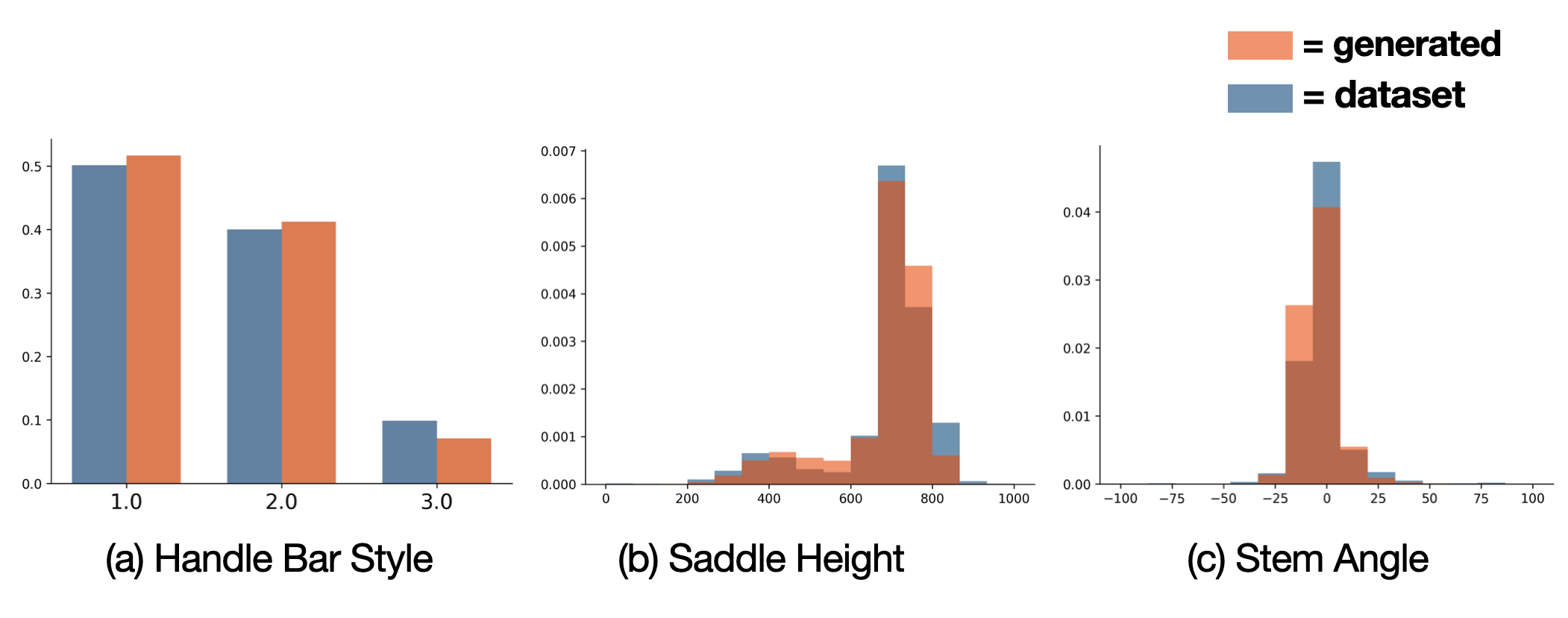}
\caption{Overall Distribution study for three selected features that were missing---handle bar style, saddle height, and stem angle}\label{fig:overall}
\end{figure}
\begin{table*}[!htbp]
\centering
\caption{Comparison of KL-Divergence Scores for different features across models. Values are scaled by $10^{-2}$ for enhanced readability. Lower scores indicate better distribution similarity. Boldface denotes the model achieving closer distribution similarity for each feature.}
\label{table:KL}
\sisetup{table-format=2.2} 
\begin{tabular}{
  l
  S
  S
  S
  S
  S
}
\toprule
{Feature} & {Saddle Height} & {Seat Tube Length} & {Stem Angle} & {Saddle Angle} & {Handlebar Angle} \\
\midrule
Ours    & \textbf{4.58} & \textbf{0.29} & \textbf{2.14} & \textbf{1.18} & 2.36 \\
TabCSDI & 6.09          & 0.49          & 6.80          & 1.49          & \textbf{1.79} \\
\bottomrule
\end{tabular}
\end{table*}
The first aspect of our analysis focuses on the overall distributions generated by the model. We test the models on the testing dataset with 10\% of features masked out and aggregate the generated results by features. Then, we calculate the KL divergence for specific features by comparing the generated population and the dataset population. We show the findings in Figure \ref{fig:overall} and Table \ref{table:KL}. From Table \ref{table:KL}, we can see our model's generated distributions are much closer to the dataset distributions for most features as measured by KL Divergence. From Figure \ref{fig:overall}, we can conclude the generated distributions are very close to those of the dataset. The findings indicate that our model excels in mirroring the dataset’s feature distributions, showcasing its ability to understand and replicate the general statistical characteristics of the engineering designs.
\subsubsection{Conditional Distributions}
\begin{figure}[t]
\centering
\includegraphics[width=8cm]{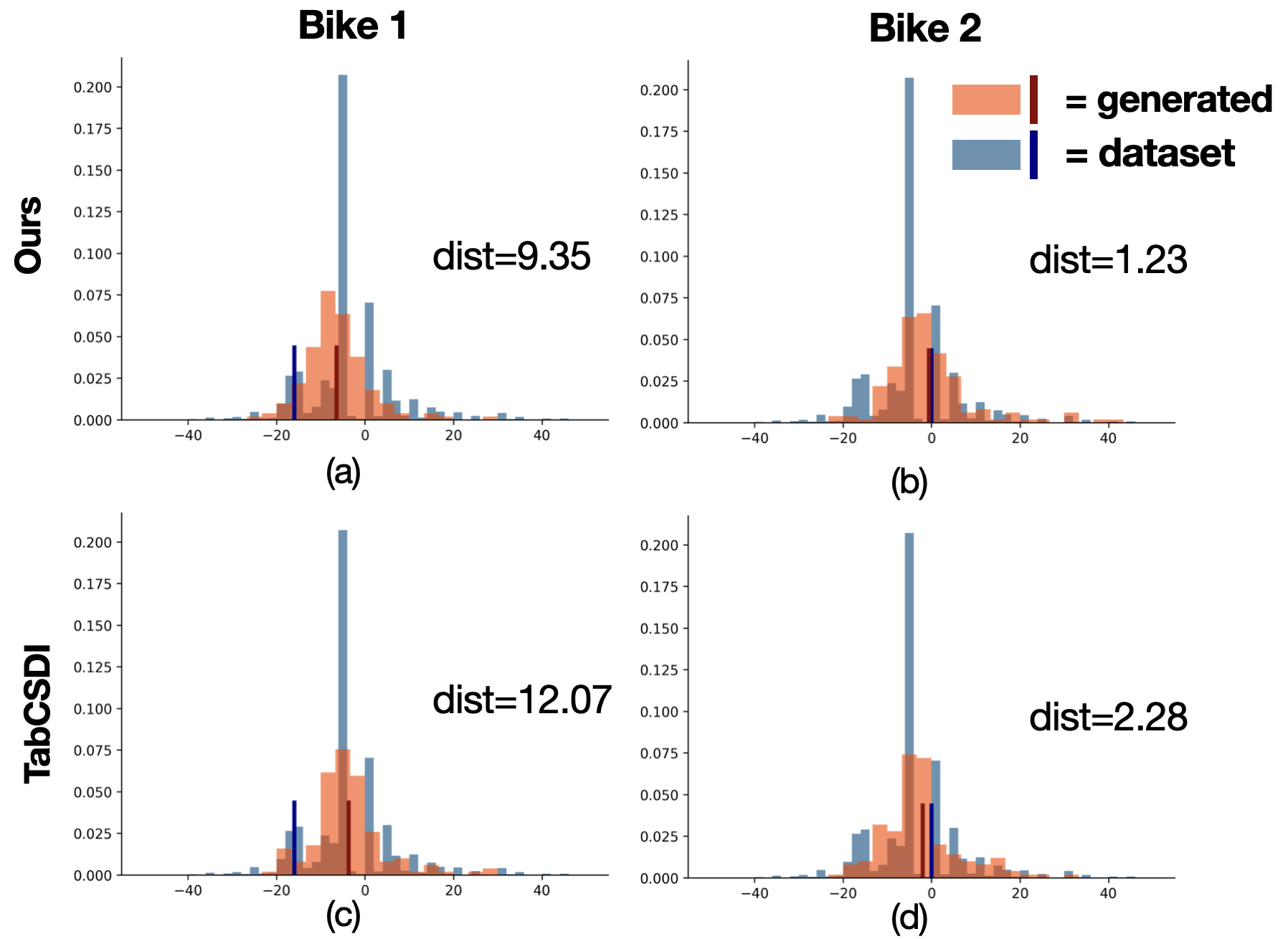}
\caption{Distribution study of feature stem angle for two randomly chosen bikes. Dark red lines represent the average value for stem angle among all the generated examples and the dark blue lines represent the dataset value for stem length for the testing bikes being studied. The "dist" number indicates the average distance between the dark red lines and the dark blue lines.}\label{fig:strongly Conditional}
\end{figure}
\begin{figure}[t]
\centering
\includegraphics[width=8cm]{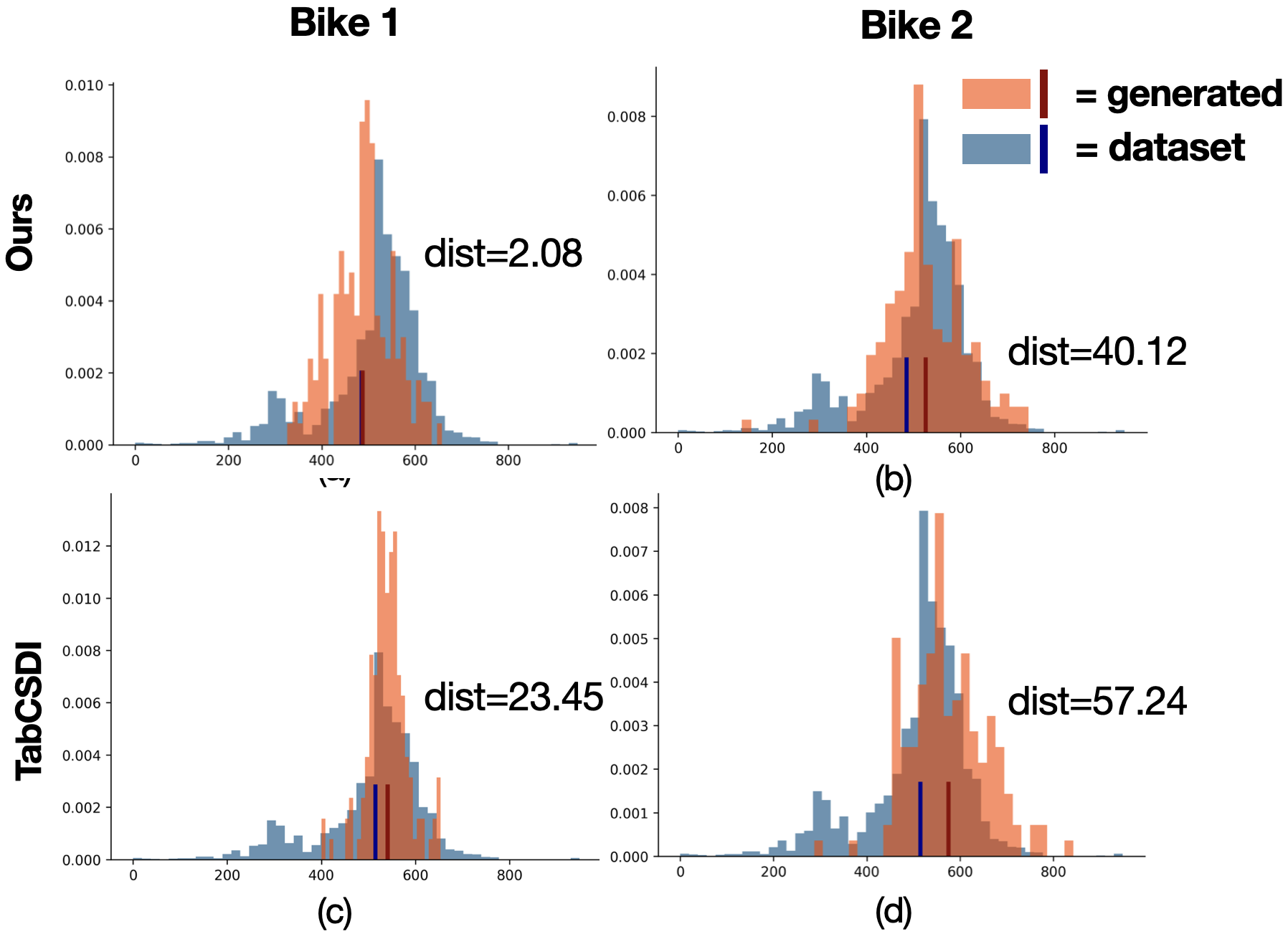}
\caption{Distribution study of feature saddle height for two randomly chosen bikes. Dark red lines represent the average value for saddle height among all the generated examples and the dark blue lines represent the dataset value for saddle height for the testing bikes being studied. The "dist" number indicates the average distance between the dark red lines and the dark blue lines.}\label{fig:relatively Conditional}
\end{figure}
\begin{figure}[t]
\centering
\includegraphics[width=7cm]{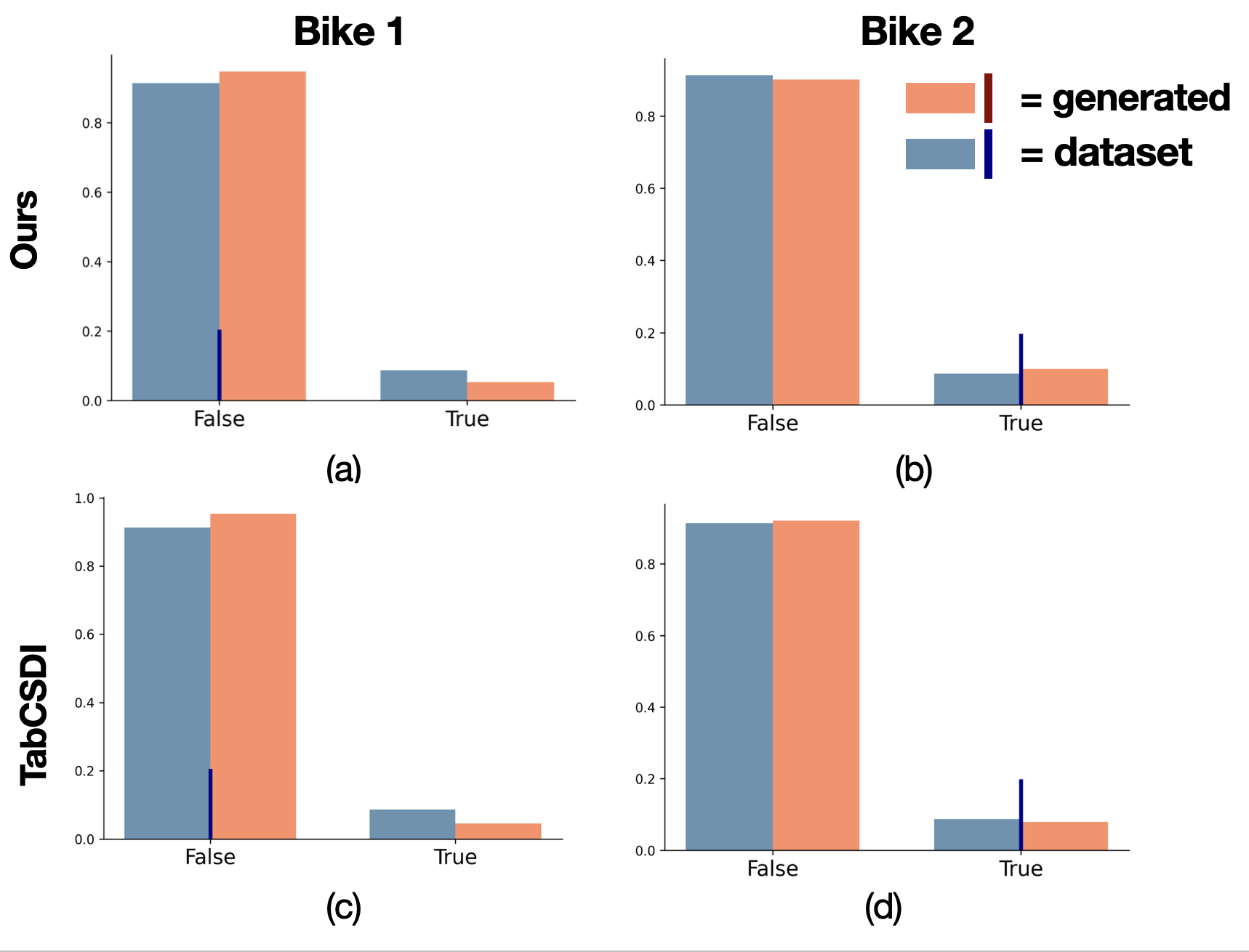}
\caption{Distribution study of feature water bottle on the down tube for two randomly chosen bikes. The dark blue lines represent the dataset value of the feature for the testing bikes being studied.}\label{fig:weakly Conditional}
\end{figure}


\begin{table*}[!htbp]
\centering
\caption{Average distance between the mean of imputed features and that of dataset values across different features, calculated as the average distance between two lines for all testing cases in Figures \ref{fig:strongly Conditional} and \ref{fig:relatively Conditional}. Bold values denote the best performance.}
\label{table:distribution}
\sisetup{table-format=1.2} 
\begin{tabular}{
  l
  S
  S
  S
}
\toprule
{Metric} & {STEM Angle} & {Seat Tube Length} & {Saddle Height} \\
\midrule
Ours     & \textbf{0.49} & \textbf{0.54} & \textbf{0.47} \\
TabCSDI  & 0.53          & 0.65           & 0.58           \\
\bottomrule
\end{tabular}
\end{table*}
Moving beyond the overall statistical landscape, we further explore the model's performance in generating features for specific bike configurations, focusing on conditional distributions. This analysis is crucial for assessing the model's precision in tailoring the generated parameters to match the unique attributes of individual designs. 
In our investigation into the model's ability to handle features with varying levels of correlation to other design parameters, we categorize the features into three distinct groups: strongly correlated, relatively correlated, and weakly correlated. This stratification allows us to assess the model's adaptability and accuracy in imputing values across different relational contexts.
\begin{enumerate}
    \item Focusing on "Stem Angle," a feature that demonstrates a strong correlation with other parameters, our analysis, illustrated in Figure \ref{fig:strongly Conditional}, delves into its distribution for two specific bike designs. The results indicate that our model not only accurately generates "Stem Angle" values but also achieves means closer to the actual data when compared to TabCSDI. This finding underscores our model's superior performance in capturing and reflecting the intricate dependencies that strongly correlated features have with other aspects of the design.
    \item Relatively Conditional: For features like "Saddle Height," which exhibit a moderate level of correlation with other features, our distribution study is presented in Figure \ref{fig:relatively Conditional}. This analysis aims to explore how our model navigates the middle ground of feature interdependencies, where the relationships are present but not as pronounced. By examining "Saddle Height," we assess the model’s ability to balance between correlation influences and independent variability within the design parameters.
    \item Weakly Correlated: For the analysis of weakly correlated features, we focus on the feature "Water Bottle on Down Tube" as a representative case. This feature, by nature, exhibits minimal correlation with other design parameters, offering an opportunity to evaluate our model's performance in scenarios where inter-feature dependencies are negligible. The results are shown in Figure \ref{fig:weakly Conditional}. The findings reveal that our model adeptly imputes values for this weakly correlated feature, maintaining fidelity to the original dataset's distribution. 
\end{enumerate}
Further, we study the average distance between mean of imputed features and that of dataset values to study the model's performance for specific features and compare with TabCSDI in Table \ref{table:distribution}. We show that our model achieves better results for all three selected features being studied.

Through this examination, we observed that the model adeptly generates features that not only align with the specificities of each bike but also reflect the nuanced conditional dependencies between different design elements. This ability to accurately impute features under varied conditional contexts demonstrates the model’s advanced understanding of the complex interrelations within the design parameters, enhancing its applicability in creating customized and contextually coherent engineering solutions.

\subsection{Autocomplete Copilot Demonstration}
In our results, we delve into the capabilities of our model as an AI copilot for engineering design, specifically focusing on its proficiency in autocompleting partial designs with a marked diversity. This exploration is segmented into three distinct cases, each highlighting the model's adeptness in generating varied and complex solutions for missing components.
\begin{figure*}[!hbt]
\centering
\includegraphics[width=15cm]{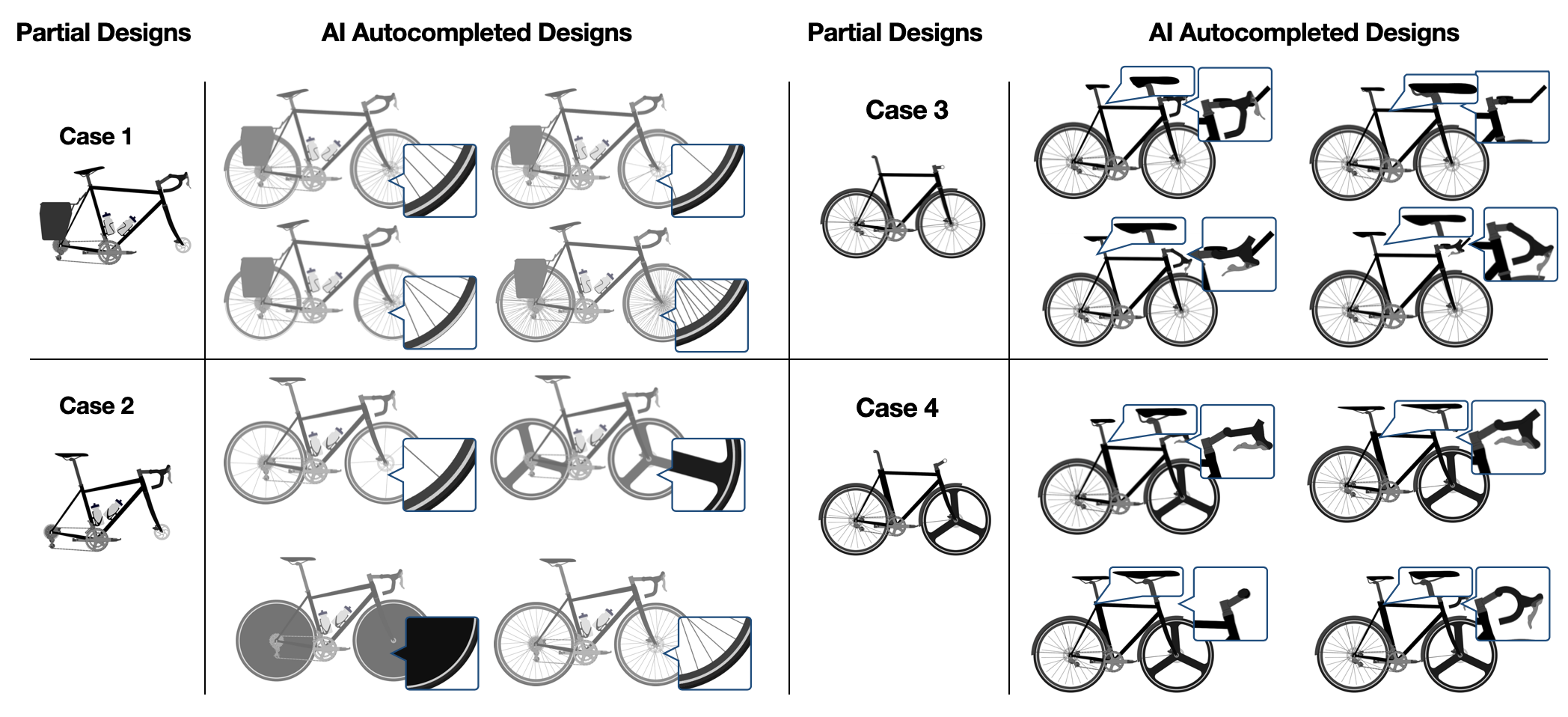}
\caption{Demo of using our model as an autocomplete copilot that generates complete designs based on different partial designs.}\label{fig:app1}
\end{figure*}
\subsubsection{Missing wheel}
When given designs lacking wheel components, our model demonstrates remarkable versatility, generating a spectrum of design recommendations that vary significantly in wheel characteristics as shown with cases 1 and 2 in Figure \ref{fig:app1}. These variations manifest in the number of spokes within the wheels, distinct differences between front and back wheel designs, and the thickness of the tire layers. 
This diversity in recommended wheel configurations is very important, as it provides a designer with varied options that they can consider for completing their design based on factors like personal preference, overall bike performance, and aesthetics. 
\subsection{Missing Handle Bar and Saddle}
In cases 3 and 4 of Figure \ref{fig:app1}, we present two scenarios where all the parameters corresponding to both the handle bar and the saddle are missing. For the handle bar, the model extends its generative capabilities to produce a wide array of handlebar designs. This includes diversity in handlebar styles, variations in brake bar angles, and adjustments in related geometrical configurations. Such detailed attention to the handlebar component not only enhances the functional adaptability of the generated designs but also their ergonomic and aesthetic appeal. Addressing designs absent of saddles, the model navigates through an extensive design space to introduce variations in saddle shapes. This capability reflects the ability of the model to propose designs that cater to different user needs and preferences while maintaining the integrity of the bike's design.


\section{Discussion and Future Work}
Our model's use of Graph Attention Neural networks informed by assembly graphs showcases a novel approach to understanding and encoding the relationships between parameters within and across different parts of an assembly design. The success in accurately imputing missing data and generating diverse, realistic designs underscores the potential of deep learning techniques in enhancing design processes. Moreover, the ability to maintain feature distribution fidelity and respect the conditional dependencies between features emphasizes the model's nuanced understanding of design parameter distribution.
One notable aspect of our findings is the diffusion model's capability to generate designs that are not only accurate but also diverse. This diversity is particularly valuable in engineering contexts where exploring a wide range of design possibilities can lead to more innovative solutions. The model's performance in generating weakly correlated features also highlights its capacity to produce coherent designs even when explicit relational cues are minimal, a testament to its ability to understand the design space and how parameters are distributed in it.

\paragraph{Limitations:}
While we show the value and efficacy of integrating generative imputation models with engineering design, this work is not without its limitations. One primary constraint is the reliance on large and well-curated datasets for the effective training of diffusion models. These models necessitate extensive data to learn the underlying distribution of design parameters accurately, which might not always be feasible or available in every engineering domain. Additionally, the necessity to define an assembly graph beforehand poses another limitation, as it requires prior knowledge of the product's structure and relationships between components. This prerequisite could limit the model's applicability in scenarios where such detailed assembly information is not readily accessible or is too complex to be encoded efficiently. Moreover, while the model demonstrates superior performance in diversity and accuracy, the computational cost associated with training and utilizing diffusion models for data imputation is significant, potentially posing challenges for real-time applications or for use on limited hardware resources. 

\paragraph{Future Work:}
Looking forward to the future, we aim to extend the capabilities of our model by incorporating real-time feedback loops that allow for an interactive exchange between the designer's inputs and the model's outputs. This could create a more dynamic and responsive design process, enabling the model to refine its suggestions based on the user's preferences and feedback. Additionally, we plan to broaden the model's application beyond bicycle design, exploring its adaptability to other engineering domains such as CAD software, automotive, or aerospace engineering. This exploration into other areas holds the potential to enhance the model's versatility as a comprehensive tool for design recommendation and completion, catering to a wide range of engineering challenges and enabling AI's role as a collaborative partner in the design process.

\section{Conclusion}
In this paper, we introduced a generative imputation model for engineering design, leveraging diffusion models and graph attention networks to accurately complete missing parametric data for design assemblies. Our model stands out by offering a solution that goes beyond traditional imputation methods, functioning as an AI design co-pilot that provides multiple viable design options for incomplete inputs, thereby facilitating a more comprehensive exploration of design possibilities. Our experimental results highlight the model's superior performance, demonstrating its ability to outperform classical imputation methods and a leading diffusion imputation model in both the accuracy of imputed values and the diversity of design options generated. Specifically, we achieved significant improvements in Root Mean Square Error (RMSE) and Error Rates for imputing missing parameters in bicycle CAD designs, alongside showcasing a marked increase in the diversity of generated designs, as quantified by our Diversity Score metric. Furthermore, through a detailed analysis of feature distributions, we showed that our model can accurately capture and reproduce the complex interdependencies between different design parameters, enabling the generation of design options that are both varied and aligned with real-world design principles. The findings of this study not only underline the potential of leveraging advanced machine learning techniques in the realm of engineering design but also open up new avenues for the application of AI as a collaborative tool in the design process. Future work will explore the extension of this model to other domains of engineering design, enhancing its capability to act as a versatile tool for design recommendation and completion across a wider range of engineering challenges.

{\small
\bibliographystyle{ieee_fullname}
\bibliography{egbib}
}
\end{document}